\documentclass[10pt,twocolumn,letterpaper]{article}

\usepackage{iccv}
\usepackage{times}
\usepackage{epsfig}
\usepackage{graphicx}
\usepackage{amsmath}
\usepackage{amssymb}

\usepackage{booktabs}
\usepackage{floatrow}
\usepackage{array}
\usepackage{makecell}
\usepackage{capt-of,calc}
\usepackage{gensymb}
\usepackage{pifont}
\usepackage{xspace}
\usepackage{arydshln}
\usepackage{multirow}
\usepackage{mathtools}
\usepackage{etoolbox}
\usepackage{courier}
\usepackage{caption}
\usepackage{subcaption}
\usepackage{amsfonts}
\usepackage{dcolumn}
\usepackage{xcolor}
\usepackage{colortbl}
\usepackage{stfloats}
\usepackage{paralist}
\usepackage{mathtools}
\usepackage{algorithm}
\usepackage{listings}
\usepackage[belowskip=0pt,aboveskip=2pt,font=small]{caption}
\usepackage[belowskip=0pt,aboveskip=2pt,font=small]{subcaption}
\usepackage[textwidth=1.8cm]{todonotes}
\usepackage[pagebackref=true,breaklinks=true,letterpaper=true,colorlinks,bookmarks=false]{hyperref}

\usepackage[capitalize]{cleveref}
\crefname{section}{Sec.}{Secs.}
\Crefname{section}{Section}{Sections}
\Crefname{table}{Table}{Tables}
\crefname{table}{Tab.}{Tabs.}

\newcommand{\objnav}{\textsc{ObjectNav}\xspace}

\newcommand{\imgnav}{\textsc{ImageNav}\xspace}
\newcommand{\xhdr}[1]{\vspace{5pt}\noindent\textbf{#1}\xspace}
\newcommand{\myquote}[1]{\textit{`#1'}}

 \def\vs{\emph{vs}\onedot}
\makeatletter
\DeclareRobustCommand\onedot{\futurelet\@let@token\@onedot}
\def\@onedot{\ifx\@let@token.\else.\null\fi\xspace}
\def\eg{\emph{e.g}\onedot} 
\def\ie{\emph{i.e}\onedot}

\newcommand*{\rowstyle}[1]{%
  \gdef\@rowstyle{#1}%
  \@rowstyle\ignorespaces%
}
\newcolumntype{=}{%
  >{\gdef\@rowstyle{}}%
}

\newcolumntype{+}{%
  >{\@rowstyle}%
}

\iccvfinalcopy %

\def\iccvPaperID{1132} %

\ificcvfinal\fi

\begin{document}
\abovedisplayskip 3.5pt plus2pt minus2pt
\belowdisplayskip \abovedisplayskip

\setlength{\abovecaptionskip}{0.5ex}
\setlength{\belowcaptionskip}{-2.0ex}
\setlength{\floatsep}{0.5ex}
\setlength{\textfloatsep}{0.5ex}

\newcommand{\csection}[1]{
    \vspace{-0.06in}
    \section{#1}
    \vspace{-0.06in}
}

\newcommand{\csubsection}[1]{
    \vspace{-0.03in}
    \subsection{#1}
    \vspace{-0.06in}
}

\newcommand{\csubsubsection}[1]{
    \vspace{-0.06in}
    \subsubsection{#1}
    \vspace{-0.05in}
}

\title{\vtwo: A simple state-of-art baseline for \imgnav and \objnav}

\author{
Karmesh Yadav$^{1}$\thanks{Equal Contribution} \,\,\,
    Arjun Majumdar$^{2}$\footnotemark[1]\,\,\,
    Ram Ramrakhya$^{2}$\,\,\,
    Naoki Yokoyama$^{2}$ \\
    Alexei Baevski$^{1}$\,\,\,
    Zsolt Kira$^{2}$\,\,\,
    Oleksandr Maksymets$^{1}$\,\,\,
    Dhruv Batra$^{1,2}$ \\
 $^1$FAIR, Meta AI \quad $^2$Georgia Institute of Technology\\
{\tt\small $^1$\{karmeshyadav,abaevski,maksymets\}@meta.com} \\
{\tt\small $^2$\{arjun.majumdar,ram.ramrakhya,nyokoyama,zkira,dbatra\}@gatech.edu} \\
{\small Code and models: \hyperlink{https://github.com/ykarmesh/OVRL}{\tt\color{black}{https://github.com/ykarmesh/OVRL}}}
}

\maketitle
\ificcvfinal\thispagestyle{empty}\fi

\begin{abstract}
We present a single neural network architecture composed of task-agnostic components (\vit{s}, convolutions, and LSTMs) that achieves state-of-art results on both the \imgnav (\myquote{go to location in $<$this picture$>$}) and \objnav (\myquote{find a chair}) tasks without any task-specific modules like object detection, segmentation, mapping, or planning modules. Such general-purpose methods offer advantages of simplicity in design, positive scaling with available compute, and versatile applicability to multiple tasks. 

Our work builds upon the recent success of self-supervised learning (SSL) for pre-training vision transformers (\vit). However, while the training recipes for convolutional networks are mature and robust, the recipes for \vit{s} are contingent and brittle, and in the case of \vit{s} for visual navigation, yet to be fully discovered. Specifically, we find that vanilla \vit{s} do not outperform ResNets on visual navigation. We propose the use of a compression layer operating over \vit patch representations to preserve spatial information along with policy training improvements. These improvements allow us to demonstrate positive scaling laws for the first time in visual navigation tasks. Consequently, our model advances state-of-the-art performance on \imgnav from 54.2\% to 82.0\% success and performs competitively against concurrent state-of-art on \objnav with success rate of 64.0\% \vs 65.0\%.

Overall, this work does not present a fundamentally new approach, but rather recommendations for training a general-purpose architecture that achieves state-of-art performance today and could serve as a strong baseline for future methods.

\end{abstract}
\vspace{-15pt}

\csection{Introduction}
\label{sec:intro}

Imagine a home assistant robot that can find things in the house. For instance, we might ask it in natural or templated language to \myquote{find a sweatshirt}. Or we may show the agent a picture of our favorite sweatshirt and ask it to find it. Designing systems for \emph{autonomous navigation to semantic goals} is a challenge of broad scientific and societal interest. 

\begin{figure}[t]
\centering
\includegraphics[width=\columnwidth]{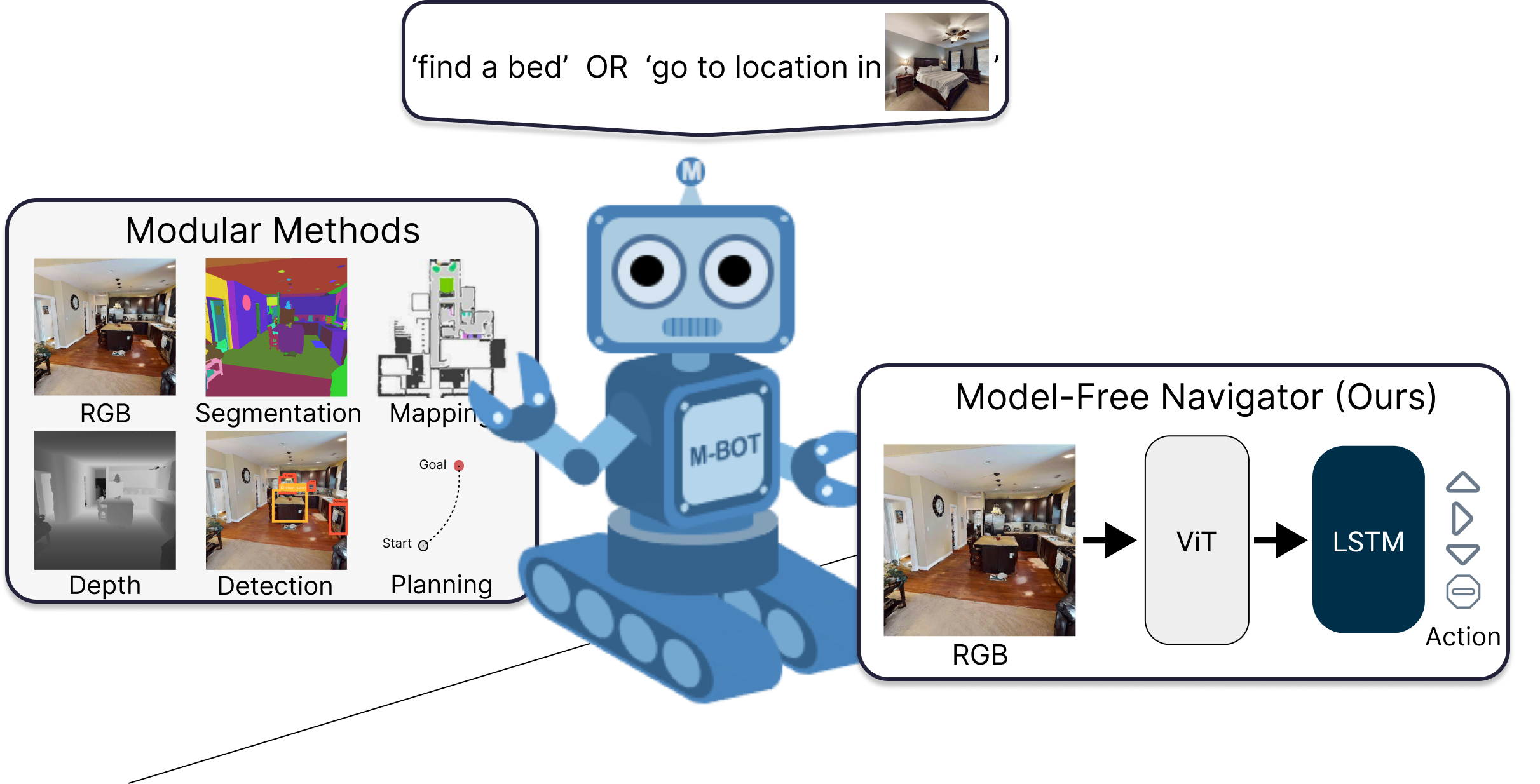}
\caption{\textbf{\vtwo} is a model-free navigator with a ViT+LSTM architecture that achieves SoTA results on \imgnav and \objnav without mapping, detectors, or segmentors of any kind.}
\label{fig:teaser}
\vspace{-1ex}
\end{figure}

In the embodied AI research community, two concrete goal specifications have emerged for semantic navigation. In \imgnav~\cite{zhu2017target}, an agent is spawned in an unseen environment and asked to find a location `described' by a goal image. In \objnav~\cite{batra2020objectnav}, it is asked to find any instance of an object category given its name \myquote{find a $<$name$>$}.

Both problems test the agent's semantic understanding and episodic memory -- what objects or parts of a scene are visible in the goal image (is it part of a kitchen or bathroom)? Where are these objects (seen in the goal image in \imgnav or mentioned by name in \objnav) typically found in a house? Where does the agent find itself at initialization? And how should it strategically search the environment to find the object or scene that it is looking for without looping over the same area multiple times?

\looseness=-1
The classical sense-plan-act pipeline~\cite{murphy_book00} from the robotics literature approaches this problem via a sequence of modules -- detecting objects (or extracting semantic features) in 2D images~\cite{maksymets2021thda,ramrakhya2022} or in 3D point clouds~\cite{wijmans2019embodied}, accumulating detections (or features) into a 2D (top-down) or 3D map~\cite{chaplot2020object,chaplot2020neural,Ramakrishnan_2022_CVPR,wasserman2022last,modular_dp} and planning a path or waypoints on this map~\cite{chaplot2020object,zhu2022navigating}, executed by a low-level controller. These approaches have been quite prevalent in prior work and have proved to be a strong baselines in both the tasks.

In this work, we advance an alternative research program -- training generalist agents constructed from task-agnostic neural components without any task-specific modules. Such general-purpose methods offer advantages of simplicity in design, positive scaling with available compute (incorporating the `bitter lesson'~\cite{sutton2019bitter}), and versatile applicability to multiple tasks. 

A flurry of recent work on image and video understanding has found that visual transformers~\cite{dosovitskiy2020image} (\vit{s}) powered by self-supervised representation learning can provide general-purpose visual representations for recognition~\cite{alexei_2022_data2vec,Chen_2021_ICCV,He_2022_mae} and generation~\cite{bao2022all,cao2022exploring} tasks. However, while the training recipes for convolutional networks are mature and robust, the recipes for \vit{s} are contingent and brittle, and in the case of \vit{s} for visual navigation, yet to be fully discovered -- and that discovery is the focus of our work.

\noindent Our key technical contributions and findings are as follows: 
\begin{asparaenum}
\item \textbf{Compression layers are needed for \vit{s} in Visual Navigation.}
We find that \vit-based agents trained from scratch perform poorly compared to \resnet{s} (e.g. achieving only 36.1\% success rate (SR) on \imgnav vs. 59.9\% for \resnet{s}). This is despite a substantially higher model capacity (\vits has $\sim$4 times more parameters than a half-width ResNet{50}). We find that a key issue with using \vit{s} for navigation problems is that both the \cls token embedding and global-average pooling remove a spatial structure that is important for the task. We propose using a compression layer (consisting of a 2D convolution plus flattening) operating over \vit patch representations to preserve spatial information, and find that it leads to \vit{s} outperforming \resnet{s} (67.4\% vs. 59.9\% SR on \imgnav).

\item \textbf{Visual pretraining unlocks positive scaling laws for the first time.}
We demonstrate, for the first time, \emph{positive scaling laws} with ViT-based agents on \imgnav. Specifically, we find that visual representation learning (using masked autoencoding (MAE)~\cite{He_2022_mae}) not only improves performance, but also enables model scaling with ViTs. With this pretraining, we are able to increase the model size from \vits to \vitb and observe gains in success rate from 80.5\% to 82.0\% (+1.5\%) and SPL (success weighted by path efficiency) from 55.2\% to 58.7\% (+3.5\%).

\item \textbf{Single architecture achieves SoTA on \imgnav and \objnav.}
Putting it all together (ViTs, compression layers, pretraining, policy training improvements and scaling), we present \textbf{\vtwo} (\emph{\vtwofull}), a simple ViT+compression-layer+LSTM architecture as a successor to the state-of-the-art method, OVRL \cite{ovrl_v1}.
\vtwo pushes the state-of-art success rate on \imgnav from 54.2\% (in \cite{ovrl_v1}) to 82.0\% (+27.8\% absolute and 51.3\% relative improvement) and on \objnav achieves 64.0\% success rate comparable to state-of-the-art (65.0\%, obtained by concurrent but orthogonal work~\cite{ramrakhya2023pirlnav}).
\vtwo agents use only RGB and GPS+Compass sensors; no egocentric depth (as used by \cite{ramrakhya2022}), no semantic segmentation (as used by \cite{ramrakhya2022}), no object detection (as used by \cite{vsm}), no semantic or geometric mapping (as used by \cite{chaplot2020object,modular_dp,Ramakrishnan_2022_CVPR,wasserman2022last,chaplot2020neural}). 
\end{asparaenum}

Overall, this work does not present a fundamentally new approach, but rather recommendations for training a general-purpose architecture that achieves state-of-art performance today and could serve as a strong baseline for future methods.

\csection{Related Work}
\label{rel_work}

\xhdr{Visual Navigation.} Visual Navigation approaches can be divided into three categories: a) SensePlanAct pipelines from classical robotics literature (typically without any learned components)~\cite{Yamauchi1997AFA}, b) Modular pipelines with learned modules~\cite{chaplot2020neural,chaplot2020object,hahn2021no,Ramakrishnan_2022_CVPR,ramrakhya2022,wasserman2022last}, and c) Monolithic neural network approaches~\cite{al2022zero,ramrakhya2023pirlnav,wijmans2019dd,ovrl_v1,ye2021auxiliary}. While many modular learning methods build explicit semantic maps~\cite{chaplot2020neural,hahn2021no}, others simply use object detectors or segmentors without mapping~\cite{ramrakhya2022,maksymets2021thda}. In comparison, we show that it is possible to achieve state-of-art performance on semantic navigation tasks without using semantic mapping, object detection, or segmentation of any kind. Such an embodied agent, composed of task-agnostic components, not only forms a strong baseline for any task but also provides the foundation for a generalist agent, which is the goal of embodied AI. 

\xhdr{ViTs in Embodied AI.} Very few works in Embodied AI have used ViTs as their vision backbone. In the Vision and Language Navigation literature,~\cite{li2022envedit} uses CLIP ViT models but keep them frozen during training. History Aware Multimodal Transformer~\cite{chen2021history} takes a ViT model pretrained on ImageNet and further pretrains the ViT along with the rest of the model using a 2 step process with proxy tasks. In comparison, \vtwo demonstrates how to finetune a SSL pretrained model using gradients from either imitation learning or reinforcement learning losses. In robotics, \cite{radosavovic2022_robot_masked} and \cite{2022_masked_wm4control} showed the benefits of using pretrained \vit but do not see any benefits of finetuning.

\xhdr{Self Supervised Learning (SSL) for Embodied AI.}
Recent works have explored self-supervised visual representations for visuomotor control~\cite{radosavovic2022_robot_masked,2022_masked_wm4control} and visual navigation~\cite{khandelwal2021simple,ovrl_v1}.
\cite{radosavovic2022_robot_masked} demonstrated the efficacy of frozen MAE representations for motor control tasks, while \cite{2022_masked_wm4control} proposed incorporating a reconstruction-based self-supervised objective alongside online model-based RL training.
EmbCLIP~\cite{khandelwal2021simple} uses off-the-shelf CLIP~\cite{radford2021learning} encoders, which are frozen during policy learning. In our work, we show the effectiveness of finetuning, not just for our method but also for ViT-based CLIP models. Both OVRL~\cite{ovrl_v1} and EmbCLIP~\cite{khandelwal2021simple} use ResNet-based visual encoders. By contrast, in this work we focus on adapting more recent ViT-based backbones for visual navigation.

\csection{Background: Tasks and Visual Pretraining}
\label{sec:background}

We study two visual navigation tasks: image-goal navigation (\imgnav)~\cite{zhu2017target} and object-goal navigation (\objnav)~\cite{batra2020objectnav}. To address these tasks, we design an embodied agent leveraging a vision transformer (\vit)~\cite{dosovitskiy2020image}. This section provides an overview of each task, then describes an approach that we use for pretraining ViTs.

\csubsection{Visual Navigation}
\label{semnav}

\begin{figure}[ht]
\vspace{-2ex}
\centering
\includegraphics[width=0.9\columnwidth]{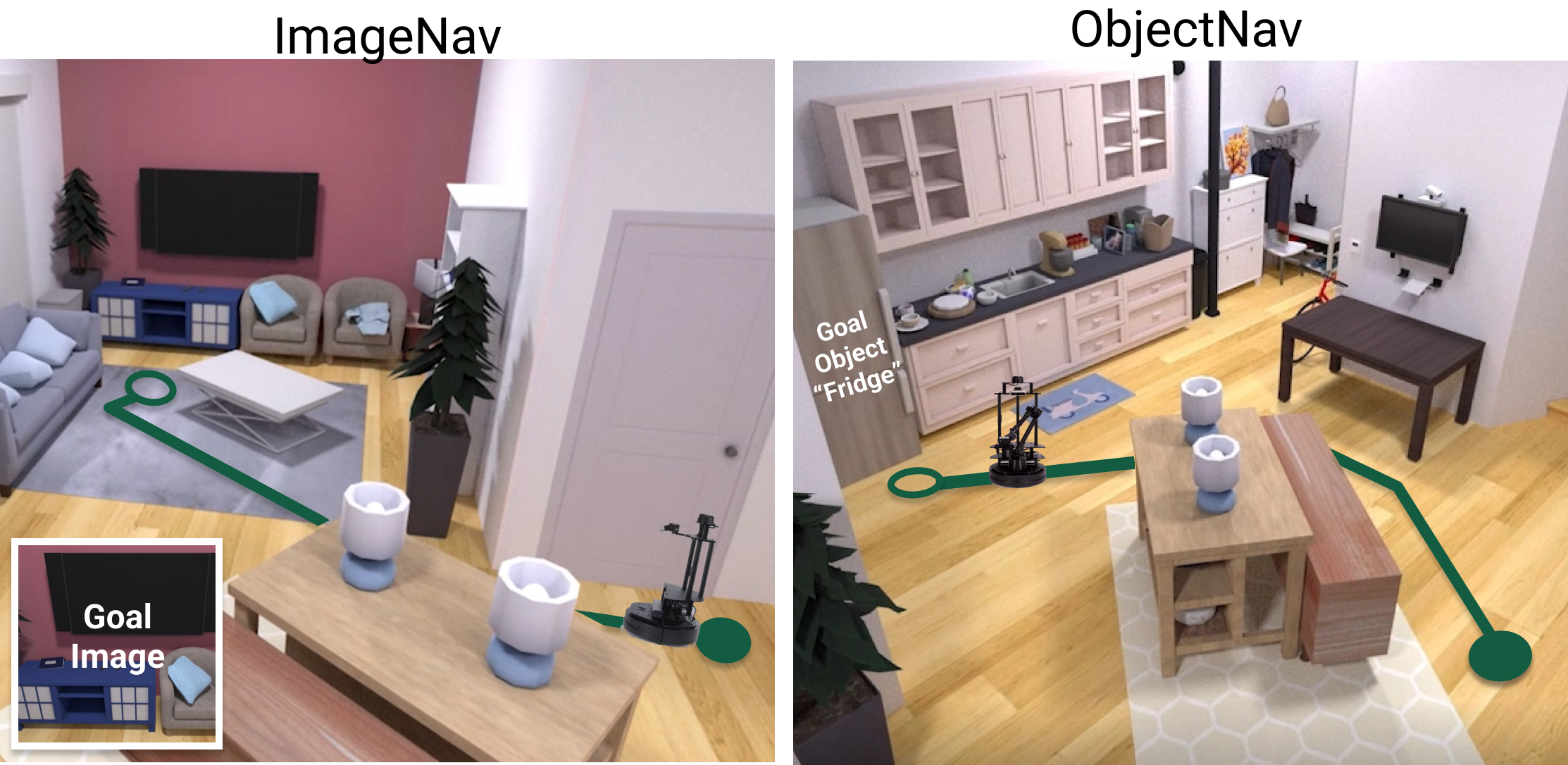}
\caption{\textbf{Visual Navigation Tasks.} In \imgnav~\cite{zhu2017target} the goal is `described' by an image and in \objnav~\cite{batra2020objectnav} the goal is described in words (e.g., \myquote{fridge}). We demonstrate the effectiveness of our `model-free navigator' (\ie, agent) on both tasks.}
\label{fig:visual_nav}
\vspace{-1ex}
\end{figure}

\cref{fig:visual_nav} illustrates the \imgnav~\cite{zhu2017target} and \objnav~\cite{batra2020objectnav} tasks. In both, an agent starts at a random position and orientation in an unknown 3D scene. The agent must explore the environment to find a goal location. In \imgnav, the goal is an image (\eg, a picture of a sofa) that is taken from the goal position. In \objnav, the agent is given the name of an object (\eg, \myquote{sofa}) that it has to find.

In these tasks, agents perceive the environment using an egocentric RGB camera. Agents navigate using a discrete action space. In \imgnav, the standard set of actions includes: \moveforward($0.25m$), \turnleft ($30^{\circ}$), \turnright ($30^{\circ}$) and \stopac to indicate that the agent thinks it has reached the goal. In \objnav, agents can also \lookup ($30^{\circ}$) and \lookdown ($30^{\circ}$).

Agents are evaluated in previously unseen environments, which allows measuring how well navigation behaviors generalize. Two standard metrics are used to assess the agent's navigation performance: success rate (SR) and success weighted by (inverse) path length (SPL)~\cite{anderson2018evaluation}. SPL rewards agents that take shorter paths to the goal, thus measuring how efficiently the agent explores new environments.

\csubsection{Masked Autoencoders (MAEs)}
\label{sec:mae}
Visual navigation tasks require understanding visual cues to navigate in new environments. Thus, agents require strong visual representations. We use masked autoencoding (MAE)~\cite{He_2022_mae} -- an efficient self-supervised visual representation learning algorithm designed for pretraining vision transformers~\cite{dosovitskiy2020image} (ViTs) -- to improve the performance of our ViT-based agent. %
MAE derives its efficiency from an asymmetric encoder-decoder design. Specifically, an input image is first divided into non-overlapping patches, a high fraction (75\%) of which are randomly masked during pretraining. The encoder only processes the remaining unmasked patches, which reduces the computational burden during pretraining. A small decoder is tasked with reconstructing the full input image. Both the encoder and decoder are ViTs, which naturally handle processing the variable number of patches. The high masking percentage is achievable due to the natural redundancy across patches in real-world images, which makes the full image predictable from only a small subset of the constituent parts. After pretraining, the decoder is discarded, and only the encoder is used for downstream tasks.

\csection{Approach}
\label{sec:approach}
\label{approach}

\begin{figure*}[t]
\vspace{-1ex}
\centering
\includegraphics[width=0.9\textwidth]{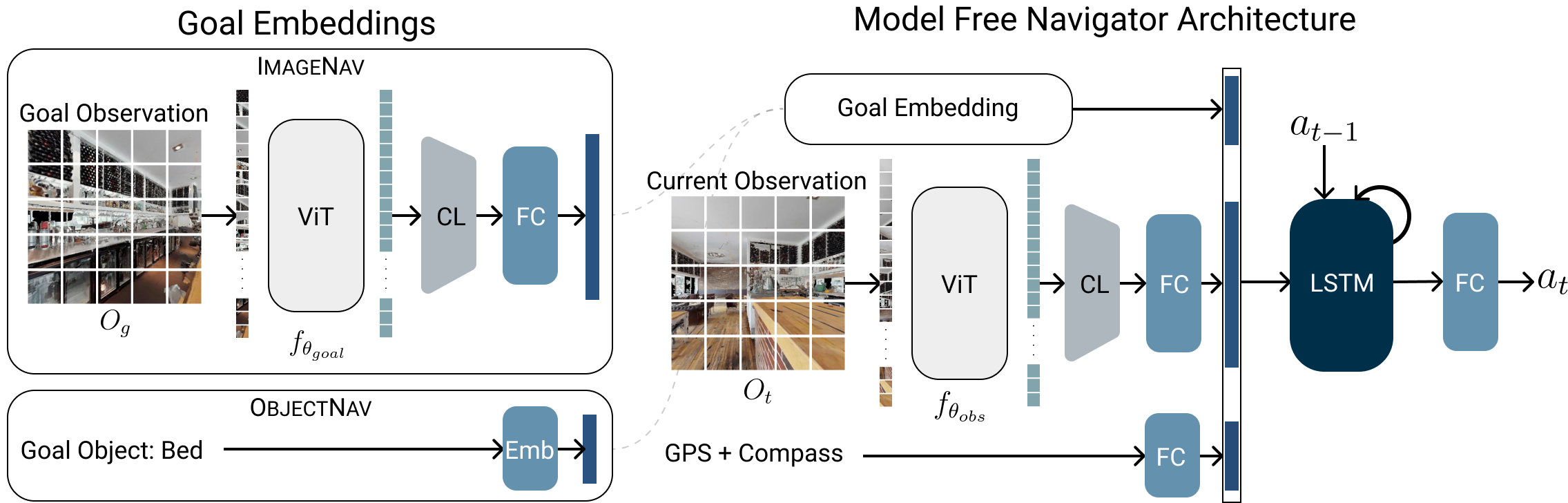}
\caption{\textbf{\vtwo architecture.} In our model-free navigator, observations $O_t$ are encoded using a from-scratch or pretrained ViT then fed to a compression layer (CL) and fully-connected layer (FC). The output representation is concatenated with a goal embedding and (optionally) a GPS+Compass encoding. Finally, an LSTM-based policy outputs actions $a_t$. In \imgnav, the visual encoder pipeline is replicated and used to encode goal images $O_g$. In \objnav, the embedding is used to encode categorical object goals (\eg, \myquote{bed})} 
\label{fig:main_fig}
\vspace{-1ex}
\end{figure*}

We use a general-purpose agent architecture for \textit{both} visual navigation tasks (\imgnav and \objnav). As shown in~\cref{fig:main_fig}, both agents primarily consist of a visual encoder (a ViT initialized randomly or pretrained with MAE), a goal encoder, and a recurrent policy network. This section describes several key components of our approach.

\begin{figure}
\vspace{-1ex}
\centering
\includegraphics[width=\columnwidth]{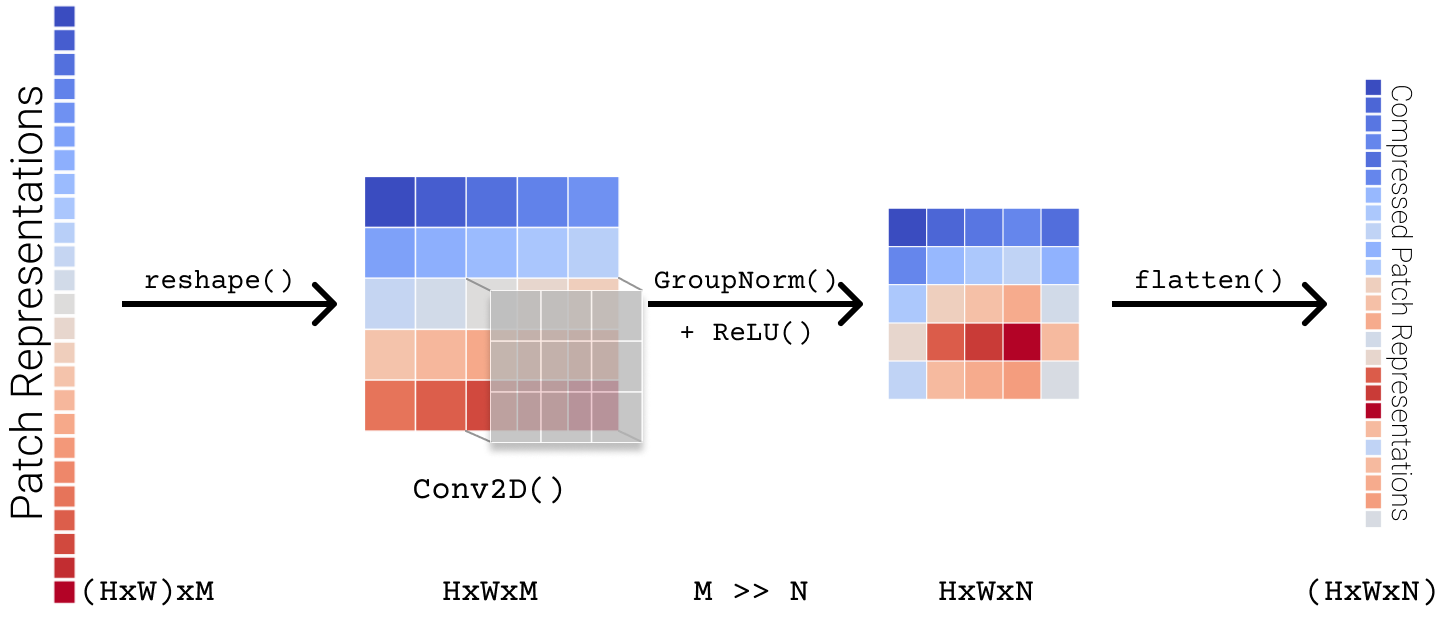}
\caption{\textbf{Compression Layer.} We propose using a compression layer to encode the output patches from a ViT encoder. The input to the compression layer are the H$\times$W output patches from ViT of size M each, where H and W are the number of patches along the height and width of the image. The patches are reshaped into a grid of size (H, W) and passed through a convolutional layer that compresses the size of the representation from M to N. The grid is then flattened and passed to the downstream model.}
\label{fig:compression_layer}
\vspace{-1ex}
\end{figure}

\label{vit_approach}
\xhdr{Compression layers for ViTs.}
As shown in~\cref{fig:main_fig}, our visual navigation agents process the RGB observation $O_t$ with a \vit-based visual encoder $f_{\theta_{obs}}$. Specifically, the input images are converted into non-overlapping 16$\times$16 patches after data augmentation, concatenated with a \cls token, and then processed with a \vit, which outputs a representation for each patch and the \cls token. In tasks such as image classification, it is common to represent the image using either (a) the \cls token output or (b) the average pooling of the patch representations (\ie, global average pooling).

Notice that both solutions remove the spatial structure present in the patch layout; we contend that this removal is detrimental for navigation. Thus, as illustrated in \cref{fig:compression_layer}, we use a layer that first reshapes the patch representations (generated by a ViT) back into a grid. Next, we process them with a convolutional layer that reduces the dimensionality of (\ie compresses) each patch.\footnote{The convolution layer is a \texttt{3$\times$3 Conv + GroupNorm + ReLU}.} Finally, we flatten the output to maintain spatially distinct features. This architecture has also been used in prior work (\eg, \cite{wijmans2019dd,ovrl_v1}) to process grid features produced by a ResNet, which we adapt here for ViTs and refer to as a \emph{compression layer}. The PyTorch implementation for this is presented in \Cref{appendix:compression_layer}.

\xhdr{Visual Navigation with ViTs}
As illustrated in~\cref{fig:main_fig}, the output from the visual encoder $f_{obs}$ is concatenated with a goal representation and an embedding of a GPS+Compass sensor (only used for \objnav) that provides pose information. The concatenated output is processed by a recurrent LSTM-based policy network, which predicts actions.

The difference between agents for each task is the method used to encode the goal.
In \imgnav, the image-goal $O_g$ is encoded with a visual encoder $f_{\theta_{goal}}$ with an identical architecture to $f_{\theta_{obs}}$. For \objnav, the goal object category (\eg \myquote{sofa}) is encoded via a learned embedding layer. 

We train our \imgnav agent with reinforcement learning (RL) using DD-PPO~\cite{wijmans2019dd} with the reward function described in a future subsection. For \objnav, we train our agent using human demonstrations with a distributed version of behavior cloning~\cite{ramrakhya2022}. Further training and evaluation details are provided in \Cref{sec:experimental_setup}.

\xhdr{Visual Encoder Pretraining.}
Our proposed approach of using a ViT-based visual encoder within a model-free navigation agent (\cref{fig:main_fig}) can be trained end-to-end from scratch (\eg, using the RL rewards described in the next section). In addition, we investigate pretraining the ViT-based visual encoder using the masked autoencoding (MAE) algorithm described in~\cref{sec:mae}. For pretraining, we collect an in-domain image dataset from HM3D~\cite{ramakrishnan2021habitat} and Gibson~\cite{xia2018gibson} scenes. This follows the observation in prior work (\eg, \cite{ovrl_v1}), which demonstrates that pretraining on in-domain data (as opposed to datasets like ImageNet) improves downstream performance. Further details about the pretraining dataset and hyperprameters are provided in~\cref{appendix:pretraining_imp}.

\xhdr{\imgnav Rewards.}
\label{sec:reward} The reward used for visual navigation is typically composed of three components: (a) a sparse reward $c_s$ for successfully completing the task, (b) a per time-step penalty $\gamma$ to incentivize efficiency, and (c) one or more reward-shaping terms to simplify the optimization problem. A common reward-shaping term is the change in (geodesic) distance to goal. Formally, let $d_t$ indicate the agent's geodesic distance to goal at time $t$; now, the reward-shaping term can be written as: $d_{t-1} - d_t$. Putting all three reward terms together, this reward is defined as:
\begin{equation} \label{eq:standard-reward}
\begin{split}
r_t = c_s &\times \bigg(\underbrace{[d_t < r_g] \And [a_t = \texttt{STOP}]}_{\mathclap{\text{\tiny{Did the agent stop close to goal?}}}} \bigg)\\ 
 &+ (d_{t-1} - d_t) - \gamma
\end{split}
\end{equation}
where $r_g$ is the goal radius and $a_t$ is the agent's action.

One limitation of the reward function in~\cref{eq:standard-reward} is that it is indifferent to the agent's `heading' at termination -- the agent is neither rewarded for looking at the goal object (which is a desirable behavior since navigation is typically a precursor to manipulation), nor is the agent penalized for ending the episode looking away from the object. To resolve this issue, \cite{al2022zero} proposed two additional angle reward terms that incentivizes, 1) turning towards the goal (using an angle-to-goal ($\theta_t$) reward shaping term) and 2) stopping while looking at the goal (using a terminal reward). Both these rewards are only awarded after the agent has entered within a goal radius $r_g$. While~\cite{al2022zero} demonstrated that their reward can improve \imgnav performance, we found that our \vtwo agent is able to hack the reward function, by never ending the episode, moving into the goal radius, turning to look at the goal, moving outside the goal radius, turning back and repeating. We provide more details about this reward and visualize the behaviour of the agent in \cref{appendix:reward_hacking_videos}. We hypothesize that prior work did not notice this exploitability because it only becomes apparent when the experiments are appropriately scaled.

We propose a principled fix to the reward function in~\cite{al2022zero}. Our key insight is that we can transform the angle-to-goal reward shaping term into a difference of potential functions, which are provably optimal for reward-shaping~\cite{ng1999policy}. Specifically, we define an angle-to-goal function $\hat{\theta}_t$ that is equal to $\pi$ outside of the goal radius and equal to the angle-to-goal otherwise:
\begin{equation} \label{eq:our-fix}
\begin{split}
\hat{\theta}_t =
\begin{cases}
\theta_t & d_t < r_g \\
\pi & d_{t} \geqslant r_g \\
\end{cases}
\end{split}
\end{equation}
With this definition, agents can be appropriately rewarded (or penalized) for entering (or exiting) the goal radius using the difference in the per timestep measure $\Delta_{\hat{\theta}_t} = \hat{\theta}_{t-1} - \hat{\theta}_t$. This term will be $0$ outside the goal radius, $\geqslant0$ when entering (thus encouraging the agent), and $\leqslant0$ when exiting (thus discouraging the agent). Furthermore, $\Delta_{\hat{\theta}_t}$ has the desirable property that any positive reward accumulated inside the goal radius will entirely be lost if the agent exits, resulting in a zero-sum path.
We take advantage of these properties, and propose the full reward structure as follows:
\begin{equation} \label{eq:our-reward}
\begin{split}
r_t = 
c_s &\times ([d_t < r_g] \And [a_t = \texttt{STOP}])\, \\ 
+ c_a &\times \bigg( \underbrace{[\theta_t < \theta_g] \And [a_t = \texttt{STOP}]}_{\mathclap{\text{\tiny{Did the agent stop facing the goal?}}}} \bigg) \\
&+ \bigg( \underbrace{\hat{\theta}_{t-1} - \hat{\theta}_t}_{\mathclap{\text{\tiny{angle-based reward shaping}}}} \bigg) +(d_{t-1} - d_t) - \gamma    
\end{split}
\end{equation}
where $\theta_g$ is an angle success threshold (set to 25$^{\circ}$ in our case). In~\cref{sec:experiments}, we find that the reward defined in~\cref{eq:our-reward} substantially improves \imgnav performance (particularly in terms of path efficiency as measured by SPL).

\csection{Experimental Findings}
\label{sec:experiments}

In this section, we first establish an \imgnav baseline that is competitive with existing SoTA methods. We then use this strong baseline to systematically address the following research questions:
\begin{asparaenum}
\item \textbf{Do ViTs work out-of-the-box for \imgnav?} No. We discover that despite higher model capacity, ViT-based agents trained from-scratch underperform smaller ResNet agents by considerable margins.
\item \textbf{How does adding a compression layer affect performance?} We find that using a compression layer to maintain spatial structure in image representations significantly improves navigation performance on \imgnav. 
\item \textbf{Does performance scale with larger ViTs?} When trained from scratch we observe mixed results. However, self-supervised visual pretraining results in consistent across-the-board improvements, along with scaling
\item \textbf{Can strong visual navigation agents `hack' the new reward function in~\cref{eq:our-reward}?} No. Agents that \textit{can} `hack' the ZER reward~\cite{al2022zero} are no longer able to `hack' the reward function with our proposed corrections.
\item \textbf{How does \vtwo performance compare with the \imgnav SoTA?} \vtwo significantly improves over prior work, including approaches that use additional cameras that provide panoramic views of the environment.
\item \textbf{Do the architectural improvements transfer to \objnav?} Yes. \vtwo outperforms \objnav SoTA in terms of SR without even using a depth sensor or segmentation module, as commonly used for \objnav.
\end{asparaenum}
\csubsection{Establishing a Strong \textbf{\imgnav} Baseline}

\begin{table}[h]
\vspace{-1ex}
\setlength{\tabcolsep}{4pt}
  \centering
  \resizebox{0.95\columnwidth}{!}{
  \begin{tabular}{clcccc}
    \toprule
    \texttt{\#} & \multicolumn{1}{c}{Visual Encoder} & Reward & \avgpool &
    SPL ($\uparrow$) & SR ($\uparrow$) \\
    \midrule
    \texttt{1} & \resnet{50} & ZER~\cite{al2022zero}     & Yes & 18.8    & 27.7          \\
    \texttt{2} & \resnet{50} & ZER~\cite{al2022zero}     & No  & 27.9    & \textbf{60.6} \\
    \texttt{3} & \resnet{50} & \cref{eq:our-reward} (ours) & No  & \textbf{33.5} & 59.9    \\
    \bottomrule
  \end{tabular}
  }
  \caption{\textbf{Our \imgnav baseline} uses a principled reward function (\cref{eq:our-reward}) and does not downsample (\avgpool) input images.}
  \label{tab:image_nav_baseline}
\vspace{-1ex}
\end{table}
\vspace{-1ex}

As a starting point, we use a baseline agent from~\cite{ovrl_v1} with a similar architecture to the model-free navigator described in~\cref{sec:approach}. Instead of the ViT-based visual encoder used in our approach, this baseline uses a half-width \resnet{50} with GroupNorm~\cite{wijmans2019dd} that is trained from scratch. We train this agent with ZER rewards~\cite{al2022zero} (\cref{eq:zer-reward}) and report results in \cref{tab:image_nav_baseline} row 1.

Next, we make two improvements to strengthen this baseline. First, we discover that removing an \avgpool operation that is used in numerous prior works (\eg,~\cite{wijmans2019dd,ovrl_v1}) to downsample images before agent processing, significantly improves performance. Specifically, removing this \avgpool operation improves \imgnav SR by +32.9\% absolute and SPL by +9.1\% (\cref{tab:image_nav_baseline} row 1 \vs 2)! We do not use this \avgpool in all the following experiments.

Finally, to avoid the potential for reward hacking (discussed in~\cref{sec:reward}), we switch to the reward function from~\cref{eq:our-reward}. In~\cref{tab:image_nav_baseline} row 3, we observe that this leads to a small (-0.7\%) drop in SR, but a large improvement in SPL of +5.6\% absolute (a +20.1\% relative improvement). Unless otherwise specified, we use the corrected reward function from~\cref{eq:our-reward} in the remaining experiments. We use this strengthened baseline from~\cref{tab:image_nav_baseline} row 3 to study the effects of switching to a ViT-based visual encoder, next. 

\csubsection{Using ViTs in a Visual Navigation Agent}
\label{sec:use_vits}
\begin{table}[h]
\vspace{-2ex}
\setlength{\tabcolsep}{4pt}
  \centering
  \resizebox{0.9\columnwidth}{!}{
  \begin{tabular}{ll cc}
    \toprule
    \texttt{\#} & \multicolumn{1}{c}{Visual Encoder} & SPL ($\uparrow$) & SR ($\uparrow$) \\
    \midrule
    \texttt{1} & \resnet{50}                    & 33.5 & 59.9 \\
    \texttt{2} & \vits \cls                  & 23.7 & 36.1 \\
    \texttt{3} & \vits Global Average Pool   & 21.1 & 35.0 \\
    \texttt{4} & \vits Compression Layer     & \textbf{37.1} & \textbf{67.4} \\
    \bottomrule
  \end{tabular}
  }
  \caption{\textbf{Compression layers} make ViTs work for \imgnav.}
  \label{tab:compression_layers}
\vspace{-1ex}
\end{table}
\vspace{-2ex}

\xhdr{Negative result.} In~\cref{tab:compression_layers} (rows 2 and 3), we switch the visual encoder in the baseline agent from a \resnet{50} to a \vits. In row 2 we use the \cls token representation produced by the ViT to represent images. In row 3, we use global average pooling over the patch representations generated by ViT. As compared with the scratch \resnet{50} baseline (row 1), we find that both solutions lead to substantially reduced performance despite the increase in model capacity as measured by number of parameters (50.9M for \vits and 21.5M for \resnet{50} agent). Specifically, the better of the two options (\cls in row 2), results in a drop of -23.8\% in SR and -9.8\% in SPL. These results indicate that, unfortunately, ViTs are not drop-in replacements for the \resnet{s} commonly used for visual navigation.

\xhdr{ViTs require Compression layer.} In~\cref{tab:compression_layers} row 4, we discover that using the compression layer described in \cref{sec:approach} reverses the negative results in rows 2 and 3, improving \imgnav SR by +7.5\% and SPL by +3.6\% over the \resnet{50} baseline in row 1. This suggests that preserving spatial structure (using a compression layer) is critical for visual navigation. Thus, we use compression layer in all remaining experiments.

\csubsection{Scaling with and without Visual Pretraining}
The encouraging performance of our \vits agent (from~\cref{tab:compression_layers} row 4, replicated in~\cref{tab:scaling_and_pretraining} row 1) suggests that further model scaling may lead to additional improvements. Unfortunately, we find that this is not the case. In~\cref{tab:scaling_and_pretraining}, we observe that simply switching from the 50.9M parameter \vits (row 1) to the 179.2M parameter \vitb (row 2) produces another negative result: SR drops -2.4\% while SPL minimally increases by +0.7\%.

\begin{table}[h!]
\setlength{\tabcolsep}{4pt}
  \centering
  \resizebox{0.8\columnwidth}{!}{
  \begin{tabular}{llc cc}
    \toprule
    \texttt{\#} & \multicolumn{1}{c}{Visual Encoder} & Pretrained & SPL ($\uparrow$) & SR ($\uparrow$) \\
    \midrule
    \texttt{1} & \vits & No & 37.1 & 67.4  \\
    \texttt{2} & \vitb & No & 37.7 & 65.0  \\
    \midrule
    \texttt{3} & \vits & Yes & 55.2 & 80.5  \\ 
    \texttt{4} & \vitb & Yes & \textbf{58.7}  & \textbf{82.0} \\ 
    \bottomrule
  \end{tabular}
  }
  \caption{\textbf{Visual pretraining} using MAE~\cite{He_2022_mae} enables positive scaling of the \vitb architectures on \imgnav.}
  \label{tab:scaling_and_pretraining}
\vspace{-1ex}
\end{table}

In contrast, we find that visual pretraining (rows 3 and 4) resolves this issue. Specifically, we pretrain ViTs using MAE (as described in~\cref{sec:background}) on the HGSP dataset (details in \cref{sec:experimental_setup}). First, with pretraining we observe a large boost in navigation performance for the \vits agent. Specifically, SR improves by +13.1\% and SPL by +18.1\% (row 1 \vs 3). Next, we find the negative scaling in rows 1 \vs 2 is reversed. With pretraining, switching from \vits to \vitb results in a +1.5\% improvement in SR and +3.5\% gain in SPL (rows 3 \vs 4) -- \ie, we finally see \emph{positive} results from model scaling. Such positive scaling was not observed in prior works such as~\cite{Xiao2022_maskedvpt_motor}.

When compared to our intial baseline, our accumulated improvements are +54.3\% in SR and +39.9\% in SPL (\cref{tab:image_nav_baseline} row 1 \vs \cref{tab:scaling_and_pretraining} row 4).

\csubsection{Comparing Reward Functions}

\begin{figure*}[t!]
\centering
\includegraphics[width=0.95\textwidth]{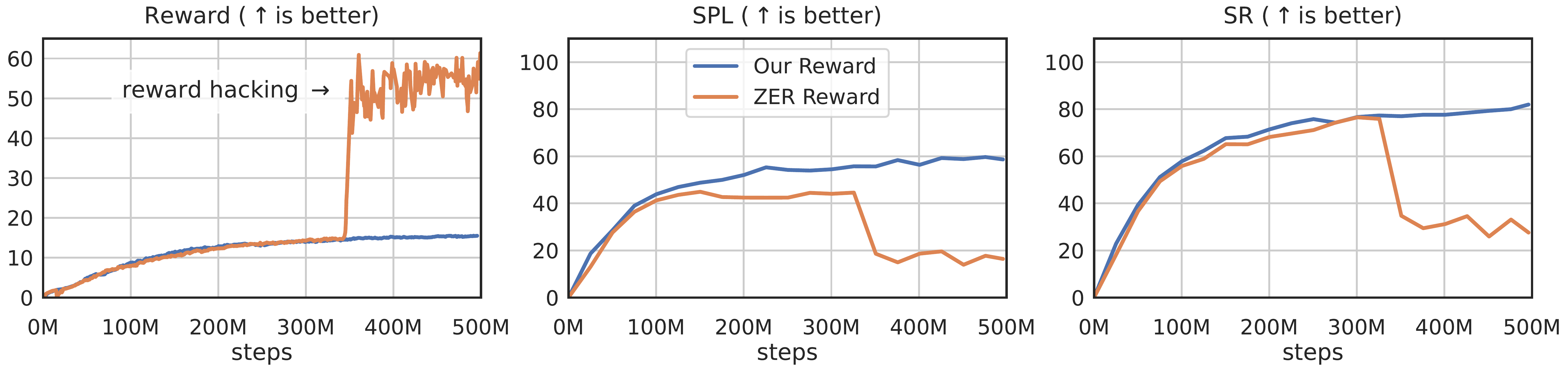}
\caption{\textbf{Reward hacking.} With the ZER reward~\cite{al2022zero} (orange curve) agents learn to hack the reward leading to large increases in training reward (left), yet substantial drops in validation path efficiency or SPL (middle) and success rate or SR (right). This undesirable behavior is resolved with the reward function introduced in~\cref{eq:our-reward} (blue curve), and performance steadily increases during training.}
\label{fig:principled_rewards}
\vspace{-1ex}
\end{figure*}

\begin{table}[h]
\vspace{-2ex}
\setlength{\tabcolsep}{4pt}
  \centering
  \resizebox{1.0\columnwidth}{!}{
  \begin{tabular}{ll cc cc}
    \toprule
    \texttt{\#} & \multicolumn{1}{c}{Visual Encoder} & Pretrained & Reward & SPL ($\uparrow$) & SR ($\uparrow$) \\
    \midrule
    \texttt{1} & \vitb & Yes & ZER\cite{al2022zero} & 44.1 & 76.5 \\
    \texttt{2} & \vitb & Yes & \cref{eq:our-reward} (ours) & \textbf{58.7} & \textbf{82.0} \\
    \bottomrule
  \end{tabular}
  }
  \caption{\textbf{Our corrected reward function} mitigates reward hacking, which leads to improved \imgnav performance.}
  \label{tab:principled_rewards}
\vspace{-1ex}
\end{table}

In~\cref{tab:principled_rewards}, we revisit the reward function used to train our visual navigation agents to confirm that using our corrected reward function (\cref{eq:our-reward}) is indeed necessary. In row 1, we find that switching back to the ZER reward (\cref{eq:zer-reward}) results in a substantial performance drop of -5.5\% in SR and a -14.6\% in SPL. In~\cref{fig:principled_rewards}, we observe that this drop is due in-part to reward hacking. Specifically, after 300M steps of training the agent using ZER rewards~\cite{al2022zero} (orange) learns to hack the reward. This corresponds to a dramatic increase in the training reward (\cref{fig:principled_rewards} left), yet precipitous drops in SPL (middle) and SR (right). Using our fix (blue), \vtwo does not hack the reward and we observe steady improvements in performance over the course of training.

\csubsection{Comparisons with the \textbf{\imgnav} SoTA}

\begin{table}[htb]
\vspace{-1ex}
\setlength{\tabcolsep}{4pt}
\renewcommand{\arraystretch}{1.0}
  \centering
  \resizebox{1.0\columnwidth}{!}{
  \begin{tabular}{=l+l+c+r+r}
    \toprule
    \texttt{\#} & Method                      & Cameras & SPL ($\uparrow$) & SR ($\uparrow$) \\
    \midrule
    \texttt{1} & ZER~\cite{al2022zero}                   & 1 Cam  & 21.6           & 29.2 \\
    \texttt{2} & ZSON~\cite{majumdar2022zson}            & 1 Cam  & 28.0           & 36.9 \\
    \texttt{3} & CRL~\cite{du2021curious}                & 1 Cam  & 10.2           & 20.4 \\
    \texttt{4} & \vone~\cite{ovrl_v1}                    & 1 Cam  & 27.0           & 54.2 \\
    \rowstyle{\color{lightgray}}
    \texttt{5} & Mem-Aug Nav~\cite{mezghani2021memory}   & 4 Cam  & 56.0           & 69.0 \\
    \midrule
    \texttt{6} & CLIP \vitb (baseline)                   & 1 Cam  & 37.4           & 51.7 \\
    \texttt{7} & \vits (baseline)                        & 1 Cam  & 37.1           & 67.4 \\
    \texttt{8} & \vtwo (ours)                            & 1 Cam  & \textbf{58.7}  & \textbf{82.0} \\
    \bottomrule
  \end{tabular}
  }
  \caption{\textbf{\imgnav performance} of our method (row 8) compared with prior SoTA (rows 1 - 5) in the single camera and 4 camera setups (in {\color{lightgray}gray}). \vtwo uses a single camera and attains better performance than methods from both the setups.}
  \label{tab:imagenav_SoTA}
\vspace{-1ex}
\end{table}

In~\cref{tab:imagenav_SoTA}, we compare \vtwo (row 8) with prior work on \imgnav (rows 1 - 5)\footnote{Details about each method is provided in \cref{appendix:baselines}.} and two baselines: a version of our full approach that uses the CLIP visual encoder (similar to EmbCLIP~\cite{khandelwal2021simple}) instead of MAE (row 6) and the \vits baseline first presented in \cref{tab:compression_layers} row 4, which uses our full method with the \vits architecture and without MAE pretraining (row 7). We observe that \vtwo outperforms all of these methods, exceeding the next best single camera (1 Cam) method, \vone (row 4) by +27.8\% in SR and +31.7\% in SPL. \vtwo also outperforms Mem-Aug Nav~\cite{mezghani2021memory}, a method that uses 4 cameras providing panoramic views of the environment and goal, by +13.0\% in SR and 2.7\% in SPL.

Additionally, we observe that the \vits baseline (row 7) is competitive with state-of-the-art methods, achieving the second highest SR and third highest SPL. Finally, the CLIP \vitb baseline (row 6) significantly underperforms \vtwo with a drop of -30.3\% in SR and -21.3\% in SPL. This indicates that pretraining with SSL on in-domain images -- and not the large-scale vision-and-language pre-training on out of domain dataset -- is a key ingredient in the success of \vtwo.

\csubsection{Transferring Improvements to \textbf{\objnav}}

\begin{table} %
\vspace{-1ex}
\setlength{\tabcolsep}{2pt}
    \centering
    \resizebox{0.95\linewidth}{!}{
        \begin{tabular}{@{}=l+l+l+r+r+c+r+r@{}}
            \toprule
            & & & \multicolumn{2}{c}{\textsc{val}} & & \multicolumn{2}{c}{\textsc{test-standard}} \\
             \cmidrule{4-5} \cmidrule{7-8}
            \texttt{\#} & Method & Camera & SPL $(\mathbf{\uparrow})$ & SR $(\mathbf{\uparrow})$
                & & SPL $(\mathbf{\uparrow})$ & SR $(\mathbf{\uparrow})$ \\
            \midrule
            \\[-10pt]
            \texttt{1} & DD-PPO~\cite{yadav2022habitat_challenge} & RGBD    & 14.2           & 27.9 & & 12.0           & 26.0 \\
            \texttt{2} & Habitat-Web~\cite{ramrakhya2022}         & RGBD-S  & 23.8           & 57.6 & & 22.0           & 55.0 \\
            \texttt{3} & Habitat-Web$^{*}$                        & RGB     & 16.0             & 41.5   & & -                & - \\
            \texttt{4} & ProcTHOR~\cite{deitke2022procthor}       & RGB     & -                & -       & & 32.0           & 54.0 \\
            \texttt{5} & Stretch~\cite{chaplot2020object}         & RGBD-S  & -                & -       & & 34.0          & 60.0 \\ 
            \texttt{6} & PIRLNav~\cite{ramrakhya2023pirlnav}      & RGB     & 34.1         & 70.4    & & 33.0           & 65.0 \\ 
            \rowstyle{\color{lightgray}}
            \texttt{7} & ByteBOT                                  & -      & -                & -        & & 37.0           & 68.0 \\ 
            \texttt{8} & \vone$^{*}$                              & RGB    & 26.8           & 62.0  & & 27.0           & 60.0 \\
            \midrule
            \texttt{9} & \vtwo (ours)                            & RGB    & 28.1           & \textbf{64.7}  & & 29.0           & 64.0 \\
            \bottomrule
            \end{tabular}
    }
    \caption{\textbf{\objnav results} on the \hmtdsem Val and Test-Standard splits. We compare to prior work on the \hmtdsem dataset and attain highest success rate on both the splits. Unpublished works that were submitted to the Test-Standard \objnav leaderboard are in {\color{lightgray}gray}. (* denotes our implementation).}
    \label{tab:objnav-leaderboard}
\vspace{-1ex}
\end{table}

This section presents a comparison of \vtwo with prior works on \objnav. In~\cref{tab:objnav-leaderboard}, we find that \vtwo achieves $64.0\%$ SR and $29.0\%$ SPL on \objnav Test-Standard split (row 9). This represents a 4.0\% improvement in SR and a 2.0\% improvement in SPL over our implementation of \vone. \vtwo is only 1\% behind PIRLNav~\cite{ramrakhya2023pirlnav} on SR, the current state-of-the-art approach. PIRLNav is a concurrent work to \vtwo which uses the pretrained \resnet encoder from \vone and learns a policy first by imitation learning (IL), followed with RL finetuning. We believe the improvements proposed by PIRLNav are orthogonal and \vtwo will further benefit from their RL finetuning strategy.

\vtwo performs better than Stretch~\cite{chaplot2020object} in terms of SR by +4.0\% (64.0\% \vs 60.0\% in row 9 \vs row 5). Stretch~\cite{chaplot2020object} uses a depth camera and an explicit semantic prediction and mapping module, and outperforms \vtwo in terms of SPL by +5\% (34.0\% \vs 29.0\% in rows 5 \vs 9). Similarly, \vtwo outperforms Habitat-Web~\cite{ramrakhya2022} (row 2), which uses an explicit semantic predictor, by +9.0\% on SR and +7.0\% on SPL (rows 2 \vs 9). We also compare to ProcTHOR~\cite{deitke2022procthor} (row 4), which uses procedural scene generation to pretrain a policy in $10k$ environments, then finetunes on \textsc{HM3D} \objnav. ProcTHOR achieves 54.0\% SR and 32.0\% SPL which is 10\% worse on SR and +3.0\% better on SPL compared to \vtwo. Our approach also improves over \vone by +4.0\% in SR and +2.0\% in SPL. We observe that the SPL is lower for all agents that learn from only human demonstrations and hypothesize that \vtwo can benefit from further RL finetuning.

Additionally, we compare \vtwo with other end-to-end methods that do not use pretrained visual encoders. First, we compare with the Habitat-Web RGB only baseline which uses a ResNet backbone and is trained using the same human demonstrations as us. We find that \vtwo is +27.7\% better on SR and +14.1\% better on SPL, demonstrating the value in using a pretrained visual encoder for IL. We also compare with a DD-PPO~\cite{yadav2022habitat_challenge} RGBD RL baseline (row 1), and find \vtwo is +36.8\% better on SR and +13.9\% better on SPL.

\begin{table*}
\footnotesize
\begin{subtable}[t]{0.32\textwidth}
\centering
\begin{tabular}[t]{l|rr}
  SSL Algorithm      & SPL ($\uparrow$) & SR ($\uparrow$) \\
  \midrule
  CLIP (Frozen)~\cite{khandelwal2021simple}   &  37.4          & 51.7          \\
  CLIP (Finetuned)  & 47.7           &  65.5         \\
  \dtv              &\textbf{59.3}   & 81.0          \\
  \mae              & 58.7           & \textbf{82.0} \\ 
\end{tabular}
\caption{\textbf{SSL algorithms} \mae and \dtv perform similarly and surpass (frozen or finetuned) CLIP.}
\label{tab:ssl_algo}
\end{subtable}%
\hspace{\fill}
\begin{subtable}[t]{0.32\textwidth}
  \centering
  \begin{tabular}[t]{l|rr}
    Representation Type & SPL ($\uparrow$) & SR ($\uparrow$) \\
    \midrule
    \cls                & 55.0            & 73.8          \\
    Global Average Pool & \textbf{59.3}            & 76.9          \\
    Compression Layer   & 58.7    & \textbf{82.0} \\ 
  \end{tabular}
   \caption{\textbf{Compression layers} work better than using the \cls token or global average pooling.}
  \label{tab:representation_style}
\end{subtable}
\hspace{\fill}
\begin{subtable}[t]{0.32\textwidth}
  \centering
  \begin{tabular}[t]{l|lrr}
    Encoder LR                 & SPL ($\uparrow$) & SR ($\uparrow$) \\
    \midrule
    2.5$\times10^{-4}$ (Default)&    37.8      & 51.0          \\
    1.5$\times10^{-6}$ (Tuned) &\textbf{58.7} & \textbf{82.0} \\
    $0.0$ (Frozen)             &    43.2      & 61.8          \\ 
  \end{tabular}
  \caption{\textbf{Learning rate tuning} substantially improves \imgnav performance.}
  \label{tab:lr}
\end{subtable}
\smallskip

\begin{subtable}[t]{0.48\textwidth}
\flushright
\centering
  \begin{tabular}[t]{l|rr}
    Augmentations & SPL ($\uparrow$) & SR ($\uparrow$) \\
    \midrule
    False         & 43.2           & 57.8          \\ 
    True          &\textbf{58.7}   & \textbf{82.0} \\ 
  \end{tabular}
  \caption{\textbf{Without augmentations} agents overfit in training.}
  \label{tab:augmentations}
\end{subtable}
 \hspace{0.02\textwidth}
\begin{subtable}[t]{0.48\textwidth}
  \centering
  \begin{tabular}[t]{l|rr}
    Pretraining Dataset  & SPL ($\uparrow$) & SR ($\uparrow$) \\
    \midrule
    OSD      & 57.5           & 81.1          \\ 
    HGSP     &\textbf{58.7}   & \textbf{82.0} \\ 
  \end{tabular}
  \caption{\textbf{Pretraining datasets} minimally impact performance.}
  \label{tab:dataset}
\end{subtable}

\caption{\textbf{Ablations} of \vtwo on \imgnav.}
\label{tab:ablations}
\vspace{-1ex}
\end{table*}

\csection{Analysis and Ablations} 
\label{analysis}
\label{sec:ablations}

This section presents ablations and a failure analysis for our \imgnav agent.

\xhdr{SSL Algorithm.}
In~\cref{tab:ssl_algo}, we use different pretraining algorithms to initialize the visual encoder. We find that \dtv~\cite{alexei_2022_data2vec} (row 3) attains similar performance to the \mae~\cite{He_2022_mae} (row 4) initialization we use in~\cref{sec:experiments}. Both methods substantially outperform a variation of \vtwo initialized with CLIP~\cite{radford2021learning} weights and then finetune (row 2) or frozen (row 1) as done in EmbCLIP~\cite{khandelwal2021simple}.

\xhdr{Using the Visual Representations.}
In \cref{tab:representation_style}, we study the impact of using a compression layer with a pretrained visual encoder (augmenting the analysis in~\cref{sec:use_vits} without pretraining). First, we find that using global average pooling (row 2) outperforms \cls token representation (row 1). Next, we observe that while global average pooling and compression layer perform similarly in terms of SPL (58.7\% \vs 59.3\%), compression layers lead to a substantially higher SR (82.0\% \vs 76.9\%). We hypothesize that preserving spatial structure with compression layers is particularly useful for recognizing the goal location (and stopping in a correct place), thus leading to a higher SR.

\xhdr{Visual Encoder Learning Rate.}
In initial experiments, we found finetuning representations pretrained with MAE or Data2Vec led to overfitting and poor generalization. Thus, we experimented with tuning the learning rate (LR) used specifically for weights of the visual encoder. In~\cref{tab:lr}, we observe that tuning the LR leads to massive improvements in SR of +31.0\% and SPL of +20.9\% (row 1 \vs 2). In fact, we find finetuning with a bad LR (row 1) is worse than simply freezing the representations (row 3).

\xhdr{Pretraining Dataset.}
In~\cref{tab:dataset} we compare the effect of pretraining with a dataset (OSD~\cite{eftekhar2021omnidata}) that was used in prior work~\cite{ovrl_v1}. We observe similar performance with both datasets, indicating that both choices are similarly `in-domain' with respect to the downstream \imgnav task.

\xhdr{Image Augmentations.} In~\cref{tab:augmentations}, we ablate the use of image augmentations during policy learning. We discover that augmentations play a vital role in preventing overfitting of the pretrained \vit agent. Without augmentations, \vtwo's SR drops by -24.2\% and SPL drops by -15.5\%.

\begin{figure}[ht]
\centering
\includegraphics[width=0.95\columnwidth]{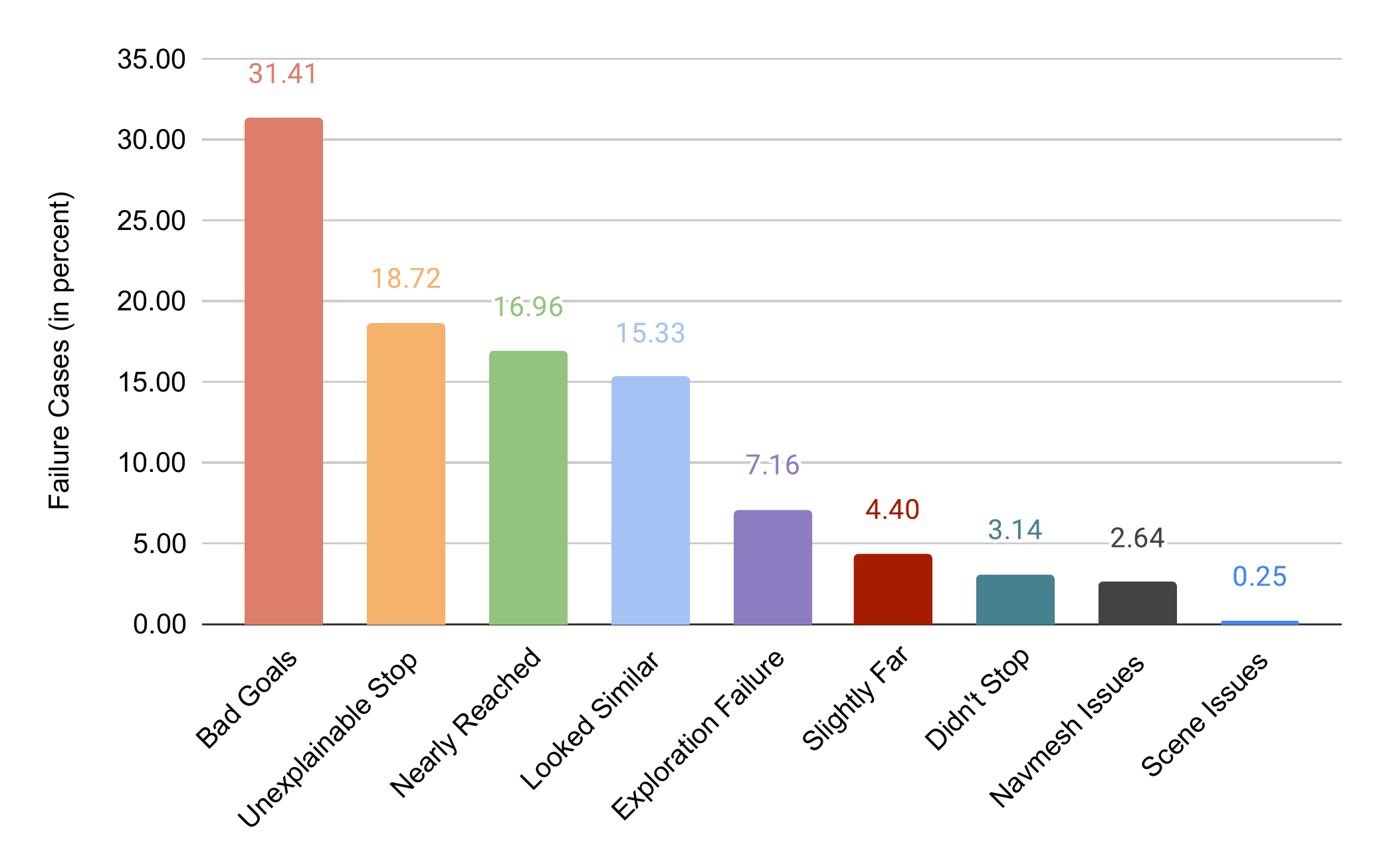}
\caption{\textbf{Breakdown of the \imgnav Failures on val split.}}
\label{fig:failures}
\vspace{-2ex}
\end{figure}

\xhdr{\imgnav Failures}
We present qualitative analysis of the failure cases of \vtwo in~\cref{fig:failures}. We perform this analysis by manually labeling each case with a cause (\eg, `exploration failure'). We find that in approximately one-third of the failures the goal-image is not semantically meaningful (`bad goals') such as blank walls. These failures account for nearly 6\% of the dataset, suggesting the upper bound on \imgnav success is 94\%. Other key issues include `unexplained stop' where the reason for stopping is unclear, `nearly reached' when the agent stops short, and `looked similar' finding a location very similar to the goal-image (but incorrect). Additional details in \Cref{appendix:failures}.

Several of these failures may be reduced with simple techniques. For example, the `nearly reached' cases may be reduced by using a stronger success criteria during training. Similarly, increasing image resolution may reduce `looked similar' failures as the agent may recognize additional visual cues to distinguish the correct goal. For other cases, such as `unexplained stop' further investigation is required.

\vspace{-1ex}
\csection{Conclusion}
\label{sec:conclusion}

In this paper, we demonstrate that a model-free navigation agent (\vtwo) composed of task-agnostic components (ViTs, convolutions, and LSTMs) can achieve state-of-the-art results on both \imgnav and \objnav. To achieve this, we show that a compression layer operating over ViT patch representations is required, which preserves the spatial information. Finally, we discover that visual pretraining with MAE enables positive scaling trends with larger ViT architectures.

\section*{Acknowledgements}

The Georgia Tech effort was supported in part by NSF, ONR YIP, and ARO PECASE. The views and conclusions contained herein are those of the authors and should not be interpreted as necessarily representing the official policies or endorsements, either expressed or implied, of the U.S. Government, or any sponsor.

{\small
\bibliographystyle{latex/ieee_fullname}
\bibliography{main}
}

\clearpage
\appendix

\section{Experimental Details}
\label{sec:experimental_setup}
\label{appendix:pretraining_imp}

\xhdr{Pretraining Dataset.} For pretraining, we create a dataset using the 800 HM3D~\cite{ramakrishnan2021habitat} and 72 Gibson~\cite{xia2018gibson} training scenes. We collect a total of 1.45M RGB images using a camera with a resolution of 512$\times$512 and a 90$^{\circ}$ FoV, attached to an oracle agent that navigates in a scene from a random start position to a random goal position using the shortest path finding algorithm. We refer to this dataset as the HM3D-Gibson Shortest Path (HGSP) dataset. We determined the HGSP dataset size (of 1.45M) based on the observation in~\cite{ovrl_v1} that pretraining on 10\% (1.45M) of the Omnidata Starter Dataset (OSD)~\cite{eftekhar2021omnidata} leads to the same downstream performance as training on 100\% of the dataset totalling 14.5M images.

\xhdr{Pretraining Details.} We pretrain visual encoders using MAE~\cite{He_2022_mae} using the same hyperparameters as~\cite{He_2022_mae}. We pretrain \vits and \vitb encoders, initialized from scratch, for 800 epochs on the HGSP dataset. We use these pretrained encoders to initialization the visual encoder for downstream task. We use the same pretrained encoder for both the \imgnav and \objnav experiments.

\xhdr{Data Augmentation}
For downstream tasks (\imgnav and \objnav), we use image augmentations during training and evaluation (\ie at test-time). Specifically, we apply random color-jitter followed by random shifts~\cite{yarats2021mastering} (\ie random translations). We apply the same image augmentation across time and to all of the samples on each GPU in our distributed training setup. For \imgnav, we use a color-jitter of 0.3 for brightness, contrast, saturation and hue levels followed by random shifts with a padding of 4 pixels. For \objnav, we use a value of 0.4 for color jitter and a padding of 16 for random shifts. These data augmentation settings follow the \imgnav experiments in~\cite{ovrl_v1}.

\xhdr{\imgnav Benchmark.} We perform \imgnav experiments using a standard dataset released by~\cite{mezghani2021memory}. This benchmark uses the Habitat simulator~\cite{savva2019habitat,szot2021habitat} and situates the task in the Gibson~\cite{xia2018gibson} environments, which include 72 training and 14 validation scenes. The validation set includes 300 episode for each scene (4,200 episodes total). In this benchmark, agents are simulated as a cylinder with a height of 1.5m, radius of 0.1m, and sensors placed 1.25m above the center of the base. The RGB camera has a resolution of 128$\times$128 and a 90$^{\circ}$ field-of-view (FoV). Agents can take up to 1000 steps in the environment, and an agent is successful if it calls \stopac within 1m of the goal position.

\xhdr{\imgnav Training Details.} We train agents in the Gibson environments for 500M timesteps (25k updates) using a total of 320 environment running in parallel. Every environment collects (up to) 64 frames of experience which is followed by 2 PPO epochs with 2 mini-batches. Unless specified explicitly, we use a learning rate of $2.5\times10^{-4}$ for training the agent and update the parameters using the AdamW optimizer~\cite{loshchilov2018decoupled} with a weight decay of $10^{-6}$. We train agents with the reward functions in~\cref{eq:zer-reward} and~\cref{eq:our-reward} from the main paper, using the following settings: success weighting $c_s = 5.0$, angle success weighting $c_a = 5.0$, goal radius $r_g = 1.0$, angle threshold $\theta_g = 25^{\circ}$, and slack penalty $\gamma = 0.01$. We evaluate performance every 25M steps of training. We report metrics based on the highest success rate (SR) achieved on the validation set.

\xhdr{\objnav Benchmark.} We conduct \objnav experiments using the \hmtdsem dataset~\cite{yadav2022hm3d}, which uses the Habitat simulator~\cite{savva2019habitat,szot2021habitat} and HM3D~\cite{ramakrishnan2021habitat} environments. The dataset consists of 80 training, 20 validation, and 20 testing scenes. We report results on the v0.1 \hmtdsem \textsc{val} and \textsc{test-std} splits that were used in the 2022 Habitat Challenge~\cite{yadav2022habitat_challenge} \objnav benchmark. In this benchmark, the agent models a LocoBot~\cite{gupta2018robot} with a height of 0.88m, radius of 0.18m, and sensors placed at the top of the agent's head. The RGB camera has a 640$\times$480 resolution and a 79$^{\circ}$ horizontal FoV. In each episode, agents must find an object drawn from one of 6 categories: \myquote{chair}, \myquote{bed}, \myquote{plant}, \myquote{toilet}, \myquote{tv/monitor}, and \myquote{sofa}. Agents are allowed 500 steps in the environment, and episodes are considered successful if the agent stops within 0.1m of a viewpoint that is (a) within 1m of any instance of an object from the goal category and (b) from which that instance is visible, following the evaluation protocol laid out in~\cite{batra2020objectnav}.

\xhdr{\objnav Human Demonstrations.} For training our imitation learning agent, we use the dataset of \objnav human demonstrations collected by Habitat-Web~\cite{ramrakhya2022,yadav2022hm3d} for \hmtdsem dataset using Amazon Mechanical Turk. The dataset consists of $77k$ human demonstrations for 80 \hmtdsem scenes that were released by~\cite{yadav2022habitat_challenge}. For each scene, we have ${\sim}158$ episodes for each unique goal object category with a randomly set start location amounting to ${\sim}950$ demonstrations per scene. The dataset amounts to a total of ${\sim}12.1$M steps in experience, with each episode averaging ${\sim}159$ steps.

\xhdr{\objnav Training Details.} We train all \objnav agents in the \textsc{HM3D} environment for ${\sim}400$M steps (25K updates) using 512 parallel environments. Similar to \imgnav, we use a weight decay of 10$^{-6}$, a learning rate of 10$^{-4}$ for the visual encoder and 10$^{-3}$ LR for everything else with the AdamW optimizer. We evaluate checkpoints after every $1000$ policy updates and report metrics for the checkpoints with the highest validation SPL.

\section{\imgnav Baselines}
\label{appendix:baselines}

This section describes the state-of-the-art \imgnav methods from prior work listed in~\cref{tab:imagenav_SoTA} (rows $1-5$).

\begin{asparaenum}
\item Zero Experience Required (\textbf{ZER})~\cite{al2022zero} proposes a novel reward function (detailed in~\cref{eq:zer-reward}) for \imgnav, which is used to train a general purpose agent composed of an ResNet-9 visual encoder (for observations and goals) and a GRU-based policy network that are trained from scratch. Our initial from-scratch ResNet-50 \imgnav baseline presented in \cref{tab:image_nav_baseline} row 1 uses a similar architecture and the same reward function as~\cite{al2022zero}.
\item Zero-Shot \objnav (\textbf{ZSON})~\cite{majumdar2022zson} uses the reward function from~\cite{al2022zero} to train an \imgnav agent consisting of a pretrained ResNet-50 encoder from~\cite{ovrl_v1} for visual observations and a CLIP~\cite{radford2021learning} visual encoder for processing goal-images. The agents in~\cite{majumdar2022zson} are trained in HM3D~\cite{ramakrishnan2021habitat} and then evaluated in Gibson~\cite{xia2018gibson}.
\item Curious Representation Learning (\textbf{CRL})~\cite{du2021curious} uses curiosity-based exploration to collect data for visual representation learning. The visual encoder is then used within a general purpose agent that is finetuned for \imgnav. We report results reproduced by~\cite{ovrl_v1}.
\item Offline Visual Representation Learning (\textbf{\vone})~\cite{ovrl_v1} pretrains a ResNet-50 encoder using the DINO~\cite{caron2021emerging} self-supervised representation learning algorithm on images from the Omnidata Starter Dataset (OSD)~\cite{eftekhar2021omnidata}. The pretrained ResNet-50 is used within a general purpose agent to process visual observations and goal images while finetuing for \imgnav with the reward function from~\cite{al2022zero}.
\item Memory-Augmented RL for \imgnav (\textbf{Mem-Aug Nav})~\cite{mezghani2021memory} enhances a general purpose agent with an attention-based memory module. In~\cite{mezghani2021memory}, agents use four RGB sensors for observations and goals, which provide a panoramic view of the environment. By contrast, our agents and the agents in~\cref{tab:imagenav_SoTA} (rows $1-4$) use a single RGB camera. Thus, we highlight results from~\cite{mezghani2021memory} in \textcolor{gray}{gray}.
\end{asparaenum}

\section{Compression Layer Implementation}
\label{appendix:compression_layer}

\begin{algorithm}[t]
\caption{PyTorch-style Compression Layer}
\label{alg:compression_layer}
\definecolor{mygreen}{rgb}{0,0.6,0}
\definecolor{mygray}{rgb}{0.5,0.5,0.5}
\definecolor{mymauve}{rgb}{0.58,0,0.82}
\lstset{
  backgroundcolor=\color{white},
  basicstyle=\footnotesize\ttfamily,
  columns=fixed,
  breaklines=true,
  commentstyle=\footnotesize\color{mygreen},
  keywordstyle=\footnotesize\color{mymauve},
}
\begin{lstlisting}[language=python]
def create_compression_layer(
    patch_dim: int,
    num_patches: int,
    approx_output_size: int = 2048,
):
    # num channels per patch
    num_channels = int(
        round(approx_output_size / num_patches)
    )

    # create layer
    layer = nn.Sequential(
        PatchReshape(),
        nn.Conv2d(
            in_channels=patch_dim,
            out_channels=num_channels,
            kernel_size=3,
            padding=1,
            bias=False,
        ),
        nn.GroupNorm(
            num_groups=1,
            num_channels=num_channels,
        ),
        nn.ReLU(inplace=True),
        nn.Flatten(),
    )

    # actual output size
    output_size = num_channels * num_patches

    return layer, output_size


class PatchReshape(nn.Module):
    def forward(self, x):
        # batch size X num patches X patch dim
        N, L, D = x.shape
        # assume square image
        H = W = int(L**0.5)
        # reshape to square grid
        x = x.reshape(N, H, W, D)
        # put channels first
        x = x.permute(0, 3, 1, 2)
        return x
\end{lstlisting}
\end{algorithm}

PyTorch-style pseudocode for creating the compression layer described in~\cref{sec:approach} is presented in~\cref{alg:compression_layer}. The layer operates on patch representations generated by a \vit-based visual encoder. It reshapes the patches into a grid and then uses a convolutional layer to compress them to a lower dimension. The dimension is determined such that when the compressed patches are subsequently concatenated (\ie, flattened) into a vector the size is approximately 2,048.

\section{\imgnav Failure Analysis Details}
\label{appendix:failures}

Here we describe each of the categories used for the \imgnav failure analysis within the main paper (which shows the distribution of these failures) in greater detail:

\xhdr{Bad Goals.} We found that a large number goals in the validation dataset face the wall or capture a noisy parts of the scene. These goals are not semantically meaningful -- \ie, they either do not `describe' a unique location in the environment or do not provide any indication of where the goal might be found.

\xhdr{Looked Similar.} The goal image and agent observation at the stopping time have some visual similarities that may have caused the agent to believe it has reached the goal. 

\xhdr{Unexplainable Stop.} The reason why the agent called stop is unclear to the annotator.

\xhdr{Nearly Reached.} The agent's distance to the goal is between 1.0m to 1.5m and the agent is looking in the direction of the goal image.

\xhdr{Slightly Far.} Distance to goal is farther than 1.5m but the agent is looking towards the goal image.

\xhdr{Exploration Failure.} The exploration strategy of the agent fails. For example, when the agent keeps looping in one small region of the environment or stop after reaching an area that is far from the goal.

\xhdr{Didn't Stop.} The agent saw the goal but did not stop.

\xhdr{Navmesh Issues.} The agent needs to pass through a narrow passage that is nearly the same size as the agent.

\xhdr{Scene Issues.} There is a hole in the scene (\ie, a missing part of the underlying 3D geometry of the environment), which confuses the agent.

\section{Additional Ablations}
\label{appendix:additional_ablations}
In this section, we present the results of additional ablations conducting on the \imgnav task with randomly initialized and pretrained model-free navigators. 

\begin{table}[h]
  \centering
  \begin{tabular}[t]{lllrr}
    \toprule
    \texttt{\#}& Pretrained & Augmentations & SPL ($\uparrow$) & SR ($\uparrow$) \\
    \midrule
    \texttt{1} & False      & False         & 31.6           & 50.0          \\ 
    \texttt{2} & False      & True          & \textbf{37.1}  & \textbf{67.4}  \\ 
    \midrule
    \texttt{3} & True       & False         & 43.2           & 57.8          \\ 
    \texttt{4} & True       & True          &\textbf{58.7}   & \textbf{82.0} \\ 
    \bottomrule
  \end{tabular}
  \caption{\textbf{Image augmentations} improve the \imgnav performance of from-scratch and pretrained agents.}
  \label{tab:scratch_augmentations}
\end{table}

\xhdr{Image Augmentations.} In~\cref{tab:scratch_augmentations}, we ablate the use of image augmentations for a randomly initialized and pretrained policy. We find that in both cases agents benefit from using augmentation during policy learning. Without pretraining, augmentations improve SR by +17.4\% and SPL by +5.5\% (rows 1 \vs 2). With pretraining, augmentations lead to gains in SR of +24.2\% and gains in SPL of +15.5\% (rows 3 \vs 4). Additionally, we find that using augmentations without pretraining (row 2) leads to a higher SR of 67.4\% than using pretraining without augmentations (row 3), which results in a SR of 57.8\%.

\begin{table}[h]
\small
  \centering
  \resizebox{1\linewidth}{!}{
  \begin{tabular}[t]{lll|rr}
    \toprule
    \texttt{\#} & Model Size             & \# Params & SPL ($\uparrow$) & SR ($\uparrow$) \\
    \midrule
    \texttt{1} & Half-width \resnet-50  & 21.6M      & 33.5             & 59.9 \\
    \texttt{2} & Full-width \resnet-50  & 59.1M      & \textbf{38.0}    & 60.0 \\
    \texttt{3} & \vits                  & 50.9M      & 37.1             & \textbf{67.4}  \\
    \bottomrule
  \end{tabular}
  }
  \caption{\textbf{Increasing ResNet-50 model size} improves SPL, but does not improve SR. Switching to \vits substantially improves SR with a minimal drop in SPL despite fewer parameters.}
  \label{tab:scratch_scaling_model}
\end{table}

\xhdr{Increasing Model Size.} The default version of our randomly initialized navigation agent uses a \vits as the vision backbone. In ~\Cref{tab:scratch_scaling_model}, we demonstrate that when trained from scratch, \vits agents are more successful than \resnet counterparts, including a full-width \resnet-50 model that uses 8.2M more parameters than the \vits variant. Specifically, we find that using a full-width \resnet-50 leads to +4.5\% improvement in SPL compared to the half-width \resnet-50. However, even with fewer parameters, \vits attains 7.4\% higher SR, while only being 0.9\% worse in terms of SPL.

\section{ZER Hacking}
\label{appendix:reward_hacking_videos}

Let $\theta_t$ denote the angle between the agent's center-of-mass and goal location and $c_a$ is angle success weighting; the ZER \cite{al2022zero} reward is written as:
\begin{equation} \label{eq:zer-reward}
\begin{split}
r_t = c_s &\times \bigg( [d_t < r_g] \And [a_t = \texttt{STOP}] \bigg) \, + \\ 
 c_a &\times \bigg( \underbrace{[\theta_t < \theta_g] \And [a_t = \texttt{STOP}]}_{\mathclap{\text{\tiny{Did the agent stop facing the goal?}}}} \bigg) \\
 &+ \bigg( \underbrace{[d_t < r_g] \times (\theta_{t-1} - \theta_t)}_{\mathclap{\text{\tiny{angle-based reward shaping}}}} \bigg) \\
 &+ (d_{t-1} - d_t) - \gamma    
\end{split}
\end{equation}
where $\theta_g$ is an angle success threshold (set to 25$^{\circ}$ in our experiments).

While~\cite{al2022zero} demonstrated that the reward in~\cref{eq:zer-reward} can improve \imgnav performance, it has a subtle but significant flaw: the reward is hackable. The culprit is that the angle-to-goal reward shaping term is not a difference of potential functions as recommended by the theory~\cite{ng1999policy}. Specifically, agents can enter the goal radius (looking away from the goal), accumulate reward by turning towards the goal, exit the goal radius, no longer face any penalty for turning around, and repeat the process.

\begin{figure}[h!]
\centering
\includegraphics[width=\columnwidth]{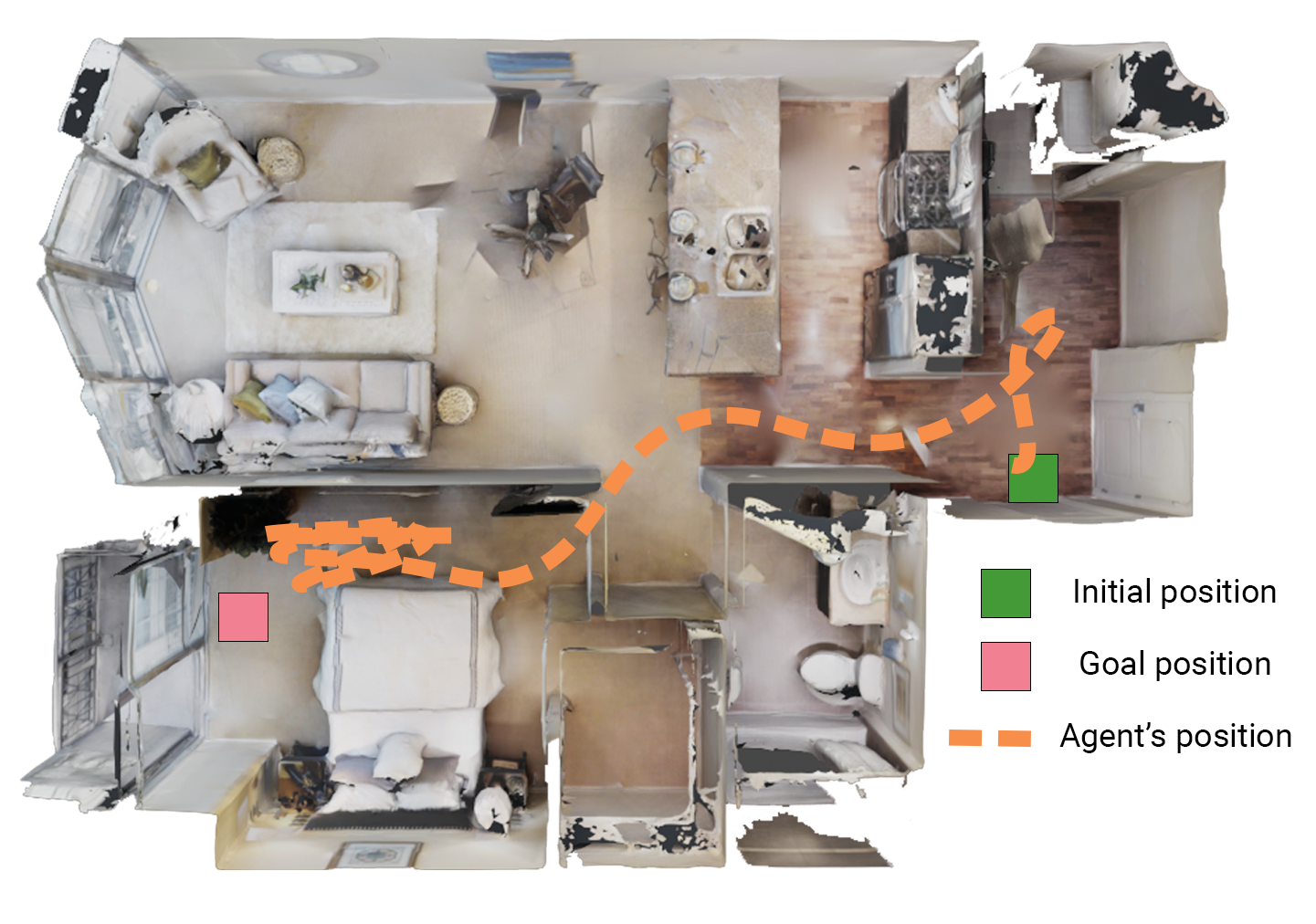}
\caption{\textbf{Reward hacking} behavior of agents trained with ZER~\cite{al2022zero} rewards. The episode's start and goal position are denoted by the green and peach square respectively. The orange dashed line indicates the agent's trajectory, which repeatedly enters and exits the goal radius (not shown) to accumulate (\ie, hack) the angle-to-goal term in the ZER~\cite{al2022zero} reward.}
\label{fig:topdown_zer_reward}
\vspace{-1ex}
\end{figure}

In \cref{fig:topdown_zer_reward}, we visualize the reward hacking behaviour of an \imgnav agent trained with the ZER~\cite{al2022zero} reward. Towards the end of the trajectory near the goal (left), the agent repeatedly enters and exits the goal radius to accumulate the angle-to-goal reward term in~\cref{eq:zer-reward} rather than successfully complete the episode.

\end{document}